\documentclass[sigconf]{acmart}

\AtBeginDocument{%
  \providecommand\BibTeX{{%
    \normalfont B\kern-0.5em{\scshape i\kern-0.25em b}\kern-0.8em\TeX}}}

\copyrightyear{2023}
\acmYear{2023}
\setcopyright{rightsretained}
\acmConference[ICAIL 2023]{Nineteenth International Conference on Artificial Intelligence and Law}{June 19--23, 2023}{Braga, Portugal}
\acmBooktitle{Nineteenth International Conference on Artificial Intelligence and Law (ICAIL 2023), June 19--23, 2023, Braga, Portugal}\acmDOI{10.1145/3594536.3595168}
\acmISBN{979-8-4007-0197-9/23/06}

\acmConference[ICAIL 2023]{Make sure to enter the correct
  conference title from your rights confirmation emai}{June 19--23, 2023}{Braga, Portugal}
\acmPrice{15.00}
\acmISBN{978-1-4503-XXXX-X/18/06}

\usepackage{xcolor}
\usepackage{graphicx}
\usepackage{pgf}
\usepackage[normalem]{ulem}
\usepackage{makecell}
\usepackage{multirow}
\useunder{\uline}{\ul}{}
\usepackage{ulem} 
\usepackage[export]{adjustbox}
\begin{document}

\title{The Perfect Victim: Computational Analysis of Judicial Attitudes towards Victims of Sexual Violence}

\author{Eliya Habba}
\affiliation{%
  \institution{The Hebrew University of Jerusalem}
  \country{Israel}
}
\email{eliya.habba@mail.huji.ac.il}

\author{Renana Keydar}
\affiliation{%
  \institution{The Hebrew University of Jerusalem}
  \country{Israel}
}
\email{renana.keydar@mail.huji.ac.il}

\author{Dan Bareket}
\affiliation{%
  \institution{Allen Institute for AI}
  \country{Israel}
}
\email{dbareket@gmail.com}

\author{Gabriel Stanovsky}
\affiliation{%
  \institution{The Hebrew University of Jerusalem}
  \institution{Allen Institute for AI}
  \country{Israel}
}
\email{gabriel.stanovsky@mail.huji.ac.il}






\begin{abstract}
  We develop computational models to analyze court statements in order to assess judicial attitudes toward victims of sexual violence in the Israeli court system. The study examines the resonance of ``rape myths'' in the criminal justice system's response to sex crimes, in particular in judicial  assessment of victim's credibility. 
We begin by formulating an ontology for evaluating judicial attitudes toward victim's credibility, with eight ordinal labels and binary categorizations. Second, we curate a manually annotated dataset for judicial assessments of victim's credibility in the Hebrew language, as well as a model that can extract credibility labels from court cases. The dataset consists of 855 verdict decision documents in sexual assault cases from 1990-2021, annotated with the help of legal experts and trained law students. The model uses a combined approach of syntactic and latent structures to find sentences that convey the judge's attitude towards the victim and classify them according to the credibility label set. Our ontology, data, and models will be made available upon request, in the hope they spur future progress in this judicial important task.
\end{abstract}






\maketitle

\section{Introduction}
Court responses to sexual violence, sexual assault and rape, have been repeatedly criticized in numerous countries and jurisdictions~\cite{temkin2018different, smith2017rape, barn2015understanding, burrowes2013responding}. Such criticisms often highlight the persistent prevalence of `rape myths' in the attitude of the criminal justice system toward victims of sex crimes, also expressed in the judicial assessment of victim's credibility.\footnote{In this work we use the term `victim' or `complainant' interchangeably when referring to survivors of sexual violence crimes (female or male), who are also testifying in court cases. We note that there are different considerations in choosing the specific terminology. While this work focuses on judicial attitudes, the potential struggles of survivors are not limited to the phase of the trial but the legal process in its entirety including police investigations and more.}

The term `rape myth' refers in general to `misled beliefs about sexual violence'~\cite{brownmiller1975against} or  `prescriptive or descriptive beliefs about rape that serve to deny, downplay or justify sexual violence'~\cite{bohner1998rape}. These myths are strongly related to wider beliefs about sex and gender~\cite{burt1980cultural} and have cultural functions that help explain their persistence in the legal system~\cite{lonsway1994rape, conaghan2014rape}. Rape myths affect how some sexual assaults are considered more ‘real’ than others~\cite{estrich1987real} and determine which assaults are taken seriously~\cite{temkin2008sexual}.

In this work we aim to assess judicial attitudes toward victims of sexual abuse in the Israeli court system, which uses the Hebrew language. To achieve this, we employ computational models to quantitatively analyze judicial statements pertaining to the assessment of the victim’s credibility in her testimony before the court. By utilizing computational models, we can analyze a large amount of data, enabling us to identify patterns and trends that would be difficult to detect otherwise.
This work is part of an on-going collaboration with The Association of Rape Crisis Centers in Israel 
(\textbf{ARCCI}).\footnote{\url{https://www.1202.org.il/en/}}


Court attitudes toward victims of sexual violence are key in understanding the criminal justice system's response to sex crimes and may help expose these implicit biases and myths. Moreover, research shows that judicial attitude toward the victim plays a substantive role in the victim's healing process~\cite{herman2005justice}.

Studies have suggested that rape myths abound in the legal system, and impact the assessment of the victim's credibility~\cite{smith2017rape}. Some examples are an on-going belief that false accusations were common, despite evidence suggesting this was unfounded~\cite{burrowes2013responding}; Juries that interpret delayed reporting or inconsistencies in a victim’s evidence as a sign of false allegations~\cite{rose2006appropriately}; Victims who are not visibly distressed, or are ‘too’ upset, are perceived as less credible~\cite{taylor2005impact}.
These studies, however, are either based on lab experiments involving mock trials and mock jurors~\cite{taylor2005impact, leverick2020we} or on a qualitative analysis of a small sample of real cases involving either judge or jury (for example, \cite{temkin2018different} examined 8 court cases).
 A computational analysis of these narratives can provide systematic insights into the judicial decision-making process in general and the court's response to the sexual assault victims in particular. 

Our contributions in this study are twofold, suggesting advances both in theoretical and empirical understanding of judicial attitudes toward victims of sexual violence as well as in computational legal analysis in a relatively low-resource language. 
First, we begin by formulating a legal ontology for the categorization of the judge's estimation of the credibility of the victim. By manually examining court cases, two legal experts in our team curated a set of 8 ordinal labels, ranging from unequivocally credible to not credible. The labels draw insights from the data while incorporating theoretical concepts from the socio-legal scholarship on rape myths and judicial assessment of victim's credibility. In addition to this fine-grained details, we also provide a coarse high-level binary categorization of these labels into ``credible'' and ``not-credible''. Future work can adapt and extend our ontology for additional study of judicial attitudes toward victims of sexual assaults in Israel as well as in other jurisdictions.
 
Our second contribution is computational, consisting of a first dataset annotating credibility in the Hebrew language, and a trained model capable of extracting these credibility labels from court cases in the Hebrew language. Towards that end, we curated a dataset of verdict decisions in sexual assault cases from the years 1990-2021. In the Israeli legal system, verdict decisions in criminal cases are decided by the trial judge and reported, in natural language, in the written decision~\cite{wenger-etal-2021-automated}. 
We collect data for 855 documents and annotate credibility labels in a portion of them with a team of legal experts and trained law students. Finally, We develop a model which takes a combined approach of both syntactic and latent structures. Specifically, we extend a syntactic algorithm (SPIKE~\cite{shlain-etal-2020-syntactic})  to Hebrew legal text to first find a cohort of sentences which convey the judge's attitude towards the victim, and then finetune a large language model to classify each sentence in the cohort according to our label set.

Taken together, we hope that the artifacts of this study (label ontology, annotated dataset, and model) will spur future work and help shed light into the narratives which inform and comprise the judicial decision-making process.

\section{Task Definition}
\begin{figure*}[t!]
    \includegraphics[scale=0.44]{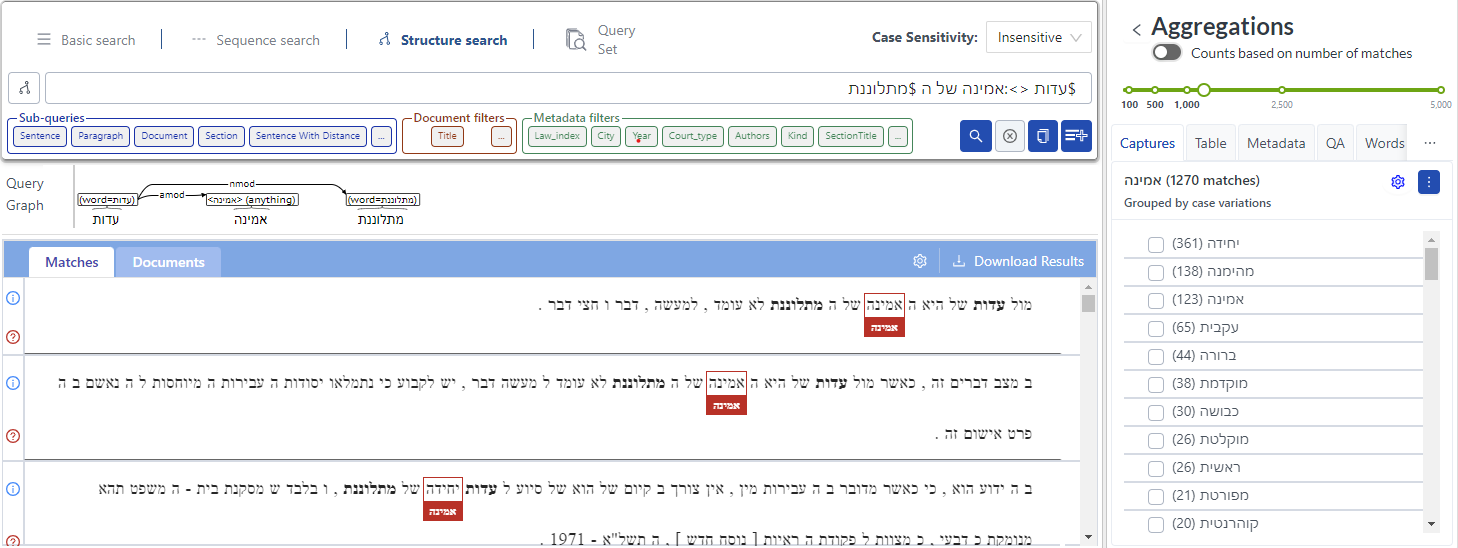}
    \centering
    \caption{SPIKE's user interface for a query matching the syntactic pattern in Figure~\ref{figure:dt}. In the top is the input text query, below it is are inferred syntactic tree and matching sentences. On the right are aggregations based on the capture provided in the query.}
    \label{figure:spike}
\end{figure*}

Given a court decision document, our task is to assess the level of credibility the judge assigns to the victim's testimony, based on an ontology we formulate in Section~\ref{section_4:legal_categories}. 
Formally, the input to the model is a document $D = (s_1, \ldots, s_n)$ composed of $n$ sentences, and the output is a set of labels $L$ from this ontology. 

Throughout this work, we follow~\cite{wenger-etal-2021-automated}, and break this overall goal into two subsequent subtasks.
First, \emph{sentence extraction} deals with extracting a list of sentences $S \subseteq D$ which convey the judge's opinion and impression of the complainant's testimony. Our underlying assumption in this task, based on domain expertise, is that each such opinion is indeed always articulated in a single sentence within the document.
Second, the \emph{credibility classification} task assigns a label 
to each of the extracted sentences.
Taken together, the concatenation of these two tasks assigns a set of labels per document.


For example, the \emph{sentence extraction} task may extract the following sentence from a court case: ``The complainant's testimony was reliable, honest, natural - sometimes she burst into tears and other times she responded in rage - and I fully trusted her'', and the assigned label by the \emph{credibility classification} task may be ``Unequivocal credibility - clear, simple, without exaggeration, authentic.''.




\section{Extractive Search in Legal Texts in Hebrew}

The {\em sentence extraction} subtask poses several challenges. It requires working with a large natural-language corpus in which the relevant sentences are linguistically diverse. Moreover, existing tools provide either no- or very-limited-support for Hebrew texts.  
The creation of an empirically-motivated annotation ontology presents similar challenges, and also requires an exploration of different overviews of the data. 
To address these challenges we develop an extractive search method based on SPIKE~\cite{shlain-etal-2020-syntactic}, an information extraction system developed for the English language.\footnote{\url{https://spike.apps.allenai.org}} 
In this section, we briefly describe SPIKE, our use of its relevant features, and how we expand and adapt it to work on Hebrew.

\subsection{Using SPIKE for Syntactic Extractive Search}
SPIKE is an extractive search system that enables users to efficiently search and extract information over a large textual corpus using syntactic patterns. SPIKE employs a query language which eliminates the need for the user to have an understanding of the underlying syntactic representations. Queries are constructed by providing a sample sentence along with simple markup, and the search is executed interactively allowing for rapid exploration, iteration, and refinement of syntax-based searches. SPIKE thus allows us to address several challenges posed by our research questions and subtasks, as defined in the previous section.

For the {\em sentence extraction} task, we devise a set of syntactic queries and run it on our corpus to retrieve a collection of relevant sentences.
Syntactic abstractions allow us to capture the diversity of phrasing used by judges without constraining the search to a fixed lexicon (see illustrations in Figure~\ref{figure:dt}). 

Devising such a query-set is a challenging task on its own, but SPIKE offers various features to aid in the process. 
First, the ``search by example'' query language does not require prior knowledge of syntactic representations. 
Second, the combination of a simple query language, an efficient graph-index and an interactive interface allows for rapid trial-and-error to refine the queries.   
Third, we use a novel {\em suggestions} method that automatically finds potentially-relevant syntactic patterns in the data based on a seed lexical search. Namely, we search the corpus for paths along the syntactic tree that connect pairs of {\em known} lexical items of interest (for example {\em ``complainant:reliable''}, {\em ``testimony:unreliable''}). We then remove some (or all) lexical constraints and keep the syntactic path. These paths potentially match relevant sentences, which contain a similar syntactic tree that is populated by other unknown lexical items. These paths are automatically translated into a SPIKE search query, ranked according to their prevalence in the corpus, and can then be reviewed and added to the query set. 

We also use SPIKE to assist the creation of our annotation ontology. To this effect, we use SPIKE's on-the-fly information-extraction and aggregation capabilities to explore different overviews of the data. 
For example, see the right hand aggregations pane in Figure~\ref{figure:spike}, which shows a list adjectives commonly used to describe the complainant's testimony. 

\subsection{Adapting SPIKE to Hebrew}
To this day, SPIKE only supported English text, and particularly those features described above. To use SPIKE on our Hebrew sentencing decision corpus, we make several additions and adaptations to the system as elaborated below.
The Hebrew version of SPIKE resulting from these development is now available online for public domain texts such as Project Ben-Yehuda.\footnote{\url{https://spike.apps.allenai.org/datasets/benyehuda}}

\paragraph{Hebrew Linguistic Annotation} this is the backbone of all core SPIKE features, responsible for automatically adding the relevant linguistic information layers (e.g. morphology, syntax, NER) to the input text. This is used both when indexing new data, and in parsing ``search by example`` search queries into a syntax-graph to search in the index. 
We created a new automatic annotation pipeline based on {\em Trankit}~\cite{nguyen-etal-2021-trankit}.\footnote{Trankit is a multilingual toolkit, which means that other languages can now be added to SPIKE with minimal effort.}
Since the English version of SPIKE already worked with Universal Dependencies annotations~\cite{nivre-etal-2017-universal}, only minor adjustments had to be made to align Trankit's output to SPIKE's format and conventions. For example, we change dependency and part-of-speech tags to canonical labels ({\em compound:smixut $\rightarrow$ compound}) and added character offsets to the sentence tokens. 

\paragraph{Hebrew-specific Heuristics} The Hebrew SPIKE version shares the UD linguistic representations with the English version, yet some phenomena remain language specific. We expanded several SPIKE features to accommodate for the Hebrew  language. 
We expand pyBART~\cite{tiktinsky-etal-2020-pybart}, an enhanced-syntactic-representation specialized for relation extraction used by SPIKE, to work with Hebrew text. We adapt existing rules to work for Hebrew by adding Hebrew terms to part-of-speech and lexical lists, and develop new Hebrew-specific rules. For example, we added new constraints that fit Hebrew's clausal complement structure to distinguish between cases such as {\em ``I want to dance''}, where an edge should be added between {\em I} and {\em dance}, and {\em ``I told him to dance''} where {\em I} and {\em dance} should not be linked.
We also add Hebrew-specific rules to SPIKE's noun-phrase expansion engine. To accommodate for Hebrew's more flexible word order, we allow a modifier to follow the noun. 

\paragraph{Query Language}  Once Hebrew was added to SPIKE, the corpus goes through the annotation pipeline, it can be queried using the existing query language as-is using Hebrew queries, which are parsed and translated to a graph-query using the same pipeline.

\paragraph{User Interface}
Multiple minor adjustments were made to properly view and edit right-to-left Hebrew text.

\section{Data Collection}

\begin{table*}[]
\begin{tabular}{ll}
\toprule
\multicolumn{1}{l}{\textbf{High Level}}                                                                                                                        & \textbf{Granular Level}                                                                                                     \\ \midrule
\multicolumn{1}{l}{\multirow{2}{*}{\begin{tabular}[l]{@{}l@{}}\textbf{Unequivocal credibility} (36.1\%)\end{tabular}}}                                               & \begin{tabular}[l]{@{}c@{}}Victim is not making fake allegations or not deemed vengeful (9.0\%)\end{tabular}               \\ 
\multicolumn{1}{l}{}                                                                                                                                           & \begin{tabular}[l]{@{}c@{}}Victim is clear, simple, without exaggeration, authentic (27.1\%)\end{tabular} \\ \midrule
\multicolumn{1}{l}{\multirow{2}{*}{\begin{tabular}[l]{@{}l@{}}\textbf{Credible because} (17.4\%)  \\ Something is \textit{strengthening} the credibility of the victim   \end{tabular}}} & \begin{tabular}[l]{@{}c@{}} Victim is credible because of external support (8.4\%)\end{tabular}                                      \\  
\multicolumn{1}{c}{}                                                                                                                                           & \begin{tabular}[l]{@{}c@{}}Victim is credible because accused is less credible (9.0\%)\end{tabular}                             \\\midrule 
\multicolumn{1}{l}{\multirow{2}{*}{\begin{tabular}[l]{@{}l@{}}\textbf{Credible but} (6.1\%)  \\ Something is \textit{undermining} the credibility of the victim \end{tabular}}}        & \begin{tabular}[l]{@{}c@{}}Victim is credible but has some problems (4.8\%)\end{tabular}                                            \\ 
\multicolumn{1}{c}{}                                                                                                                                           & \begin{tabular}[l]{@{}c@{}}Victim is credible but it is a sole testimony or lacks external support  (1.3\%)\end{tabular}                  \\ \midrule
\multicolumn{1}{l}{\multirow{2}{*}{\textbf{Not credible} (12.9\%)}}                                                                                                     & \begin{tabular}[l]{@{}c@{}}Victim presented several versions or was not consistent (9.0\%)\end{tabular}                     \\ 
\multicolumn{1}{c}{}                                                                                                                                           & \begin{tabular}[l]{@{}c@{}}Victim is deemed vengeful, not authentic, or making fake allegations (3.9\%)\end{tabular}                        \\ \midrule
\multicolumn{2}{l}{\begin{tabular}[l]{@{}l@{}}\textbf{Not relevant} (27.4\%)\\ Irrelevant to the judge's stance on the victim's credibility \end{tabular}}                                                                                                                                                                                                          \\ \bottomrule
\end{tabular}
\caption{Hierarchical structure of the credibility labels.  The percentage next to each category represents its proportion in the gold dataset relative to other labels within the same level of categorization in the tagged data.}
\label{table:train_label_distribution}
\end{table*}

In this section, we describe the process of creating the dataset, starting with the collection of a corpus of documents, selecting relevant sentences, and labeling a portion of it. 
Our data collection process is depicted in Section~\ref{sec:data_collection} where we compile a corpus of verdict decisions from Israel Magistrate and District Courts, as well as sentences relating to the judicial assessment of the victim's credibility. In Section~\ref{sec:ontology}, we establish an ontology for the narrative explanation of credibility, based on a mixed-methods computational grounded theory.
Specifically, we utilized unsupervised clustering of the sentences extracted using SPIKE to group similar sentences. Based on these groups, we defined categories and iteratively refined them using legal experts research on the impact of rape myths in literature. This iterative process ultimately led to the final set of granular categories, which were then used to classify each selected sentence into one of the legal categories presented in Table~\ref{table:train_label_distribution}.

A description of the annotation guidelines is given in Section~\ref{sec:annotation_guidelines}, which is used to classify each selected sentence into one of the legal categories. Section~\ref{sec:annotation_interface} describes our annotation interface, addressing the legal complexities of the task and allowing for a controlled  annotation process. Section~\ref{sec:inter_annotator_agreement} focuses on the inter-annotator agreement, calculated using overlapping sentences annotated by multiple annotators. Finally, The results of the annotation task are presented in Section~\ref{sec:label _distribution}, including a breakdown of the annotated sentences across the categories.

\subsection{Data Collection}
\label{sec:data_collection}
We compiled a corpus containing 855 verdict decisions from Israel Magistrate and District Courts, from the years 1988 - 2021, as collected by Nevo legal database.\footnote{\url{https://www.nevo.co.il}. The data does not represent all the cases that were held in court but only those that were documented in the Nevo database.} All the cases in the corpus deal with sexual offenses under sections 345-351 of the Israel Penal Law, 5737-1977, including offenses of rape, sodomy, indecent acts and sex offenses within the family. Due to the sensitivity of the materials, this corpus will only be made available upon request.
Using SPIKE, we identified, for each verdict decision, the sentences relating the judicial assessment of the victim's credibility for classification based on the ontology we will establish.  

\subsection{An Ontology for Judicial Assessment of Victim's Credibility}
\label{sec:ontology}
The first step in assessing judicial attitudes toward victims was to establish an ontology for narrative explanation of credibility. We used a mixed-method approach of computational grounded theory designed to allow categories and themes to emerge inductively from data, while informing our interpretation of these themes based on theoretical understandings of the underlying social world~\cite{nelson2020computational}. 

\label{section_4:legal_categories}
Applied to our case, we used a clustering model that grouped similar sentences together. Based on the patterns and relationships between these grouped sentences, and informed by the existing literature on rape myths in the criminal justice system, we identified key categories for the ontology, on two levels of analysis: high level and granular level, as shown in Table~\ref{table:train_label_distribution}.


On the basis of these categories, we devised an annotation scheme to ensure the accuracy and consistency of our data. Using the annotation scheme, each sentence was categorized into the most appropriate legal category.

\subsection{Annotation Guidelines} 
\label{sec:annotation_guidelines}
In the annotation task, annotators were asked to classify each selected sentence from the verdict decision into one of 
eight
categories pertaining to the level of credibility of the victim's testimony as assessed by the judge. 
Annotators were required to use the granular-level categories, as each category at the granular level corresponds to exactly one high-level category.  This approach allowed for efficient and accurate tagging of the data, contributing to the quality of the annotation process.

Assigning categories to judicial assessments of credibility is a complex legal task. Judicial statements can be ambiguous, contain specialized legal jargon, or require an understanding of the context. Additionally, there may be a lack of consensus among legal experts on the exact  classification of a given statement. All of these factors contribute to the complexity of the annotation task and require careful consideration and attention during the annotation process. 

To ensure the accuracy of the annotation process and the quality of its results, we assembled a team of legal experts and trained law students and, each tasked with annotating on average 80 sentences, with 25 sentences overlapping between annotators to measure inter-annotator agreement. 

\begin{figure}[tb!]
  \includegraphics[]{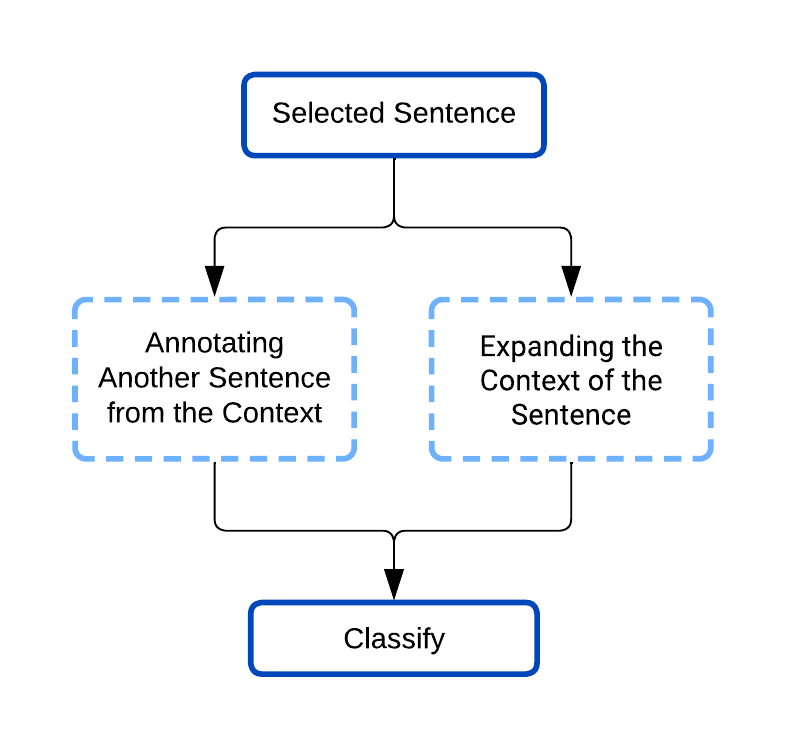}
  \caption{Annotation interface for sentence tagging and correction. The dotted squares indicate optional actions for the tagger, which may not be necessary for every tagging instance.}
  \label{figure:interface_first_part}
\end{figure}

\subsection{Annotation Interface} 
\label{sec:annotation_interface}
\begin{figure}[tb!]
  \includegraphics[scale=0.55]{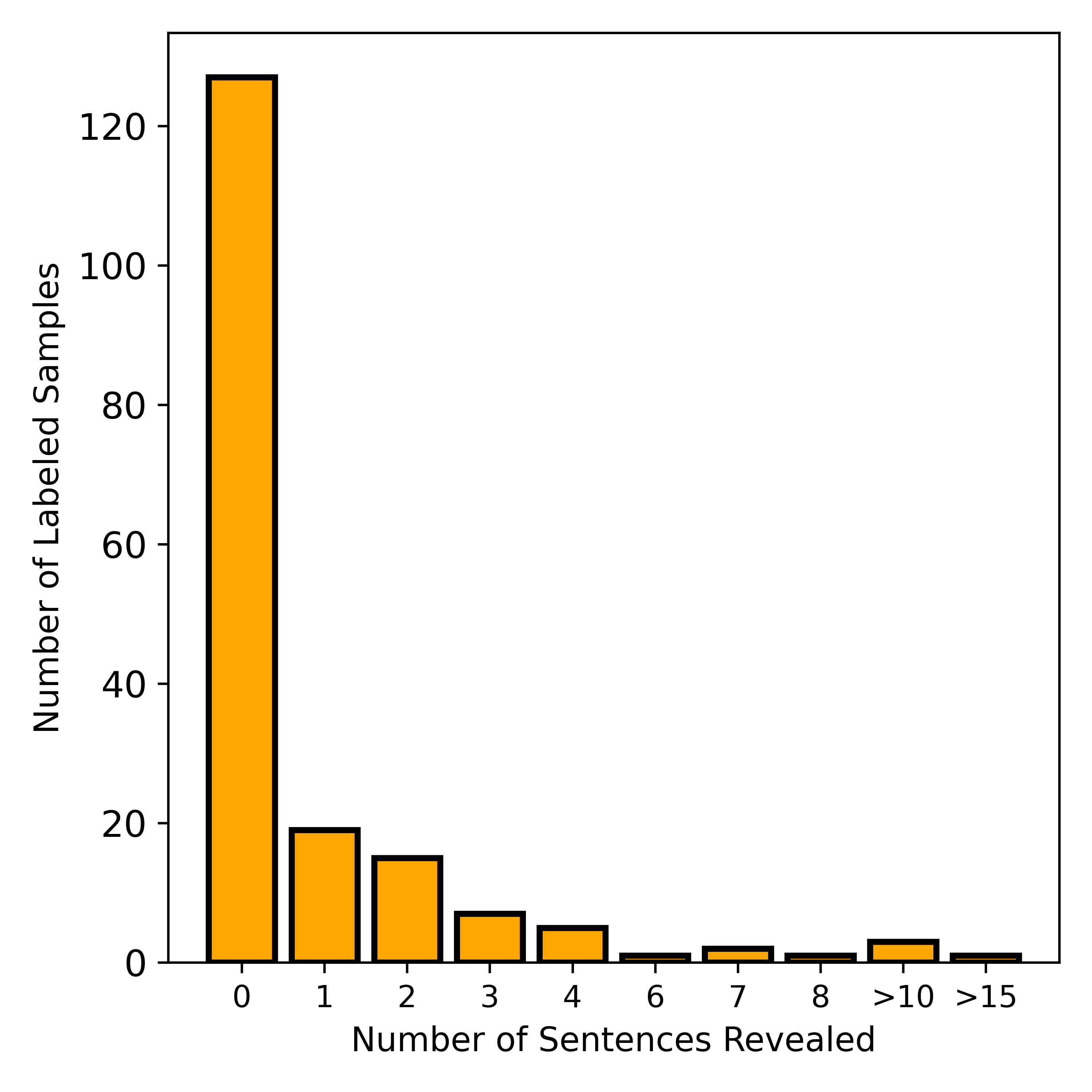}
  \caption{The distribution of the additional sentences before and after the main sentence required for determining the label by the annotators. In most cases only the main sentence was needed for accurate classification.}
  \label{figure:len_of_context}
\end{figure}

Our annotation interface, as depicted in Figure~\ref{figure:interface_first_part}, was designed to address the legal complexities of the task at hand. The interface we developed allows for a controlled and informed annotation process. Specifically, it displays the sentence selected by SPIKE from the verdict decision containing the judicial assessment of victim's credibility. The annotator can then choose whether to classify the sentence to one of the granular level categories shown in Table~\ref{table:train_label_distribution} or use control buttons, to expand the context window to view previous and next sentences. The annotators were instructed to reveal more sentences only if they could not classify the selected sentence  without the additional context. Moreover, they were instructed to reveal the minimal number of additional sentences necessary for classification. The design of our annotation scheme is also meant to limit the length of text that annotators had to read, thus minimizing their exposure to distressing descriptions, unless absolutely necessary.  
The annotation interface  tracks the additional sentences viewed by each annotator, which allow us to determine the amount of context needed to accurately classify a judicial statement as shown in Figure~\ref{figure:len_of_context}.

Alongside the classification to categories, annotators can also tag a sentence as not relevant in case of a sentence that does not relate the judge's impression of the victim's testimony. This allows the annotator to identify sentences that were extracted incorrectly and helps  improve our data. By marking a sentence as not relevant, the annotator can then select another sentence from the window context which is the correct sentence that should have been extracted and add it to the list of sentences to be annotated. This method helps improve the accuracy of sentence extraction in documents where errors were present. 44\% of sentences initially annotated as not relevant were easily corrected using the Annotation Interface and added to the list of tagged sentences, despite not being extracted through the sentence extraction process.\\

Given the complexity of the task and the ambiguity inherent to language, annotators could classify a sentence into a secondary category, without assigning significance to the order of the two tagged categories.


\subsection{Inter-Annotator Agreement} 
\label{sec:inter_annotator_agreement}
To validate the results of the annotation task, it is important to ensure the consistency of annotations. For that, we randomly selected 25 sentences and asked all five annotators to classify them. To evaluate the agreement between the annotators, we used Krippendorff's alpha, considering at both the high level and the granular level categories, and found that the agreement on the high level categories was fairly good, with a score of 0.7, if we exclude the not relevant category.

Figure~\ref{figure:confusion_matrix} shows inter-annotator agreement. The labeling of one human annotator was compared with the majority agreement of all annotators on shared sentences to determine the accuracy of human annotators. The average accuracy was found to be 0.72 for high-level labels and 0.54 for granular-level labels. These results indicate that the task of labeling is not straightforward even for human annotators and that there may not always be agreement between different annotators on the granular-level labels. This highlights the importance of using high-level labels as well, to achieve a wider agreement on the labeling.\\ 
While a variety of reasons led to disagreements between annotators, two recurring issues were identified. In some instances, the judge was impressed by the investigation conducted to the complainant. However, the annotators disagreed on whether this was due to the trustworthiness of the complainant or external factors such as the thoroughness of the investigation.

For example: ``After reviewing the transcripts of the complainant's investigation, and watching the CD documenting the investigation, the complainant's testimony left a positive impression on me.'' 

In other cases, the judge concluded that it was impossible to convict because of  difficulties in the complainant's testimony, but the annotators disagreed on the judge's impression of the testimony. The annotators were divided in their assessment of whether the judge viewed the complainant as untrustworthy, credible but with limitations, or credible but unable to secure a conviction.
For example,
 ``Since the complainant's testimony is the only testimony, this doubt is added to the doubts that arose in her testimony, and these led me to the conclusion that it is impossible to base incriminating findings on the basis of the complainant's sole testimony regarding the offenses attributed to the accused''. 

\begin{figure}[tb!]
\centering
\input{images/confusion_matrix.pgf}
\caption{Comparison of accuracy between each tagger and the majority agreement of all taggers on two levels: high level and granular level.  This comparison highlights the complexity of the task, as it is challenging to reach a consensus on granular labels. However, the agreement of ~70\% show a higher agreement on the high-level labels.} 
\label{figure:confusion_matrix}
\end{figure}
\subsection{Label Distribution Analysis} 
\label{sec:label _distribution}The distribution of labels in our dataset is presented in Table~\ref{table:train_label_distribution}. We find that the most commonly occurring label is that of  \textsf{``unequivocal credibility''}. This conforms to our expectations given that more than 60\%  of sex assault cases which reach the court end in a plea bargain and only a minority of the cases culminate a verdict.\footnote{https://www.1202.org.il/centers-union/publications/reports/662-2021-annual-report; https://www.gov.il/BlobFolder/generalpage/prkfiles2/he/2021-year-report.pdf} 
Nevertheless, even when the judicial assessment of the victim's testimony is positive, we see that in 9\% of the instances, this is because the victim did not make the impression of being vengeful or falsely alleging on the accused. This reasoning, while seemingly supporting a ``positive'' judicial attitude, echoes some of the rape myths discussed above, mainly the unsupported belief regarding the prevalence of false accusations. This resonates with the finding that in 13\% of judicial assessments, the testimony of the victim was found to be \textsf{``Not credible''}. The negative evaluation of the victim was mostly because of lack of consistency, but also due to fear of false accusations and vengeful motives.

These findings highlight, once again, the lingering presence of rape myths regarding false accusations. They also substantiate what~\cite{temkin2008sexual} termed as 
``the justice gap'' in sexual assault cases, referring to the dramatic gap between victimization and conviction. Part of the justice gap has to do with the discrepancy between the judicial expectation for consistency rooted in  formal rules of evidence and the unique characteristics of sexual assault victimization~\cite{ellison2005closing, hohl2015complaints}.
The second most frequent label is that of \textsf{``Not relevant''} with 103 instances. We find that these instances were often quotes from the parties' briefs rather than direct assertion by the judge. Only a small number of sentences were found to be completely unrelated to the complainant's testimony. These findings highlight the complexity and nuances of the task and further emphasize the importance of clear annotation guidelines and a thorough understanding of the context.



\section{Models}
In this section, we describe the methods we use to extract relevant sentences and classify them into predefined legal categories. Section~\ref{sec:sent_ext} outlines the sentence extraction task, which includes the use of our Hebrew-adapted SPIKE, while Section~\ref{sec:classification} discusses the credibility classification task. 

\subsection{Modeling Sentences Extraction}
\label{sec:sent_ext}
\begin{figure}[tb!]
\centering
    \includegraphics[scale=0.4,trim={0 0 8cm 1.5cm}]{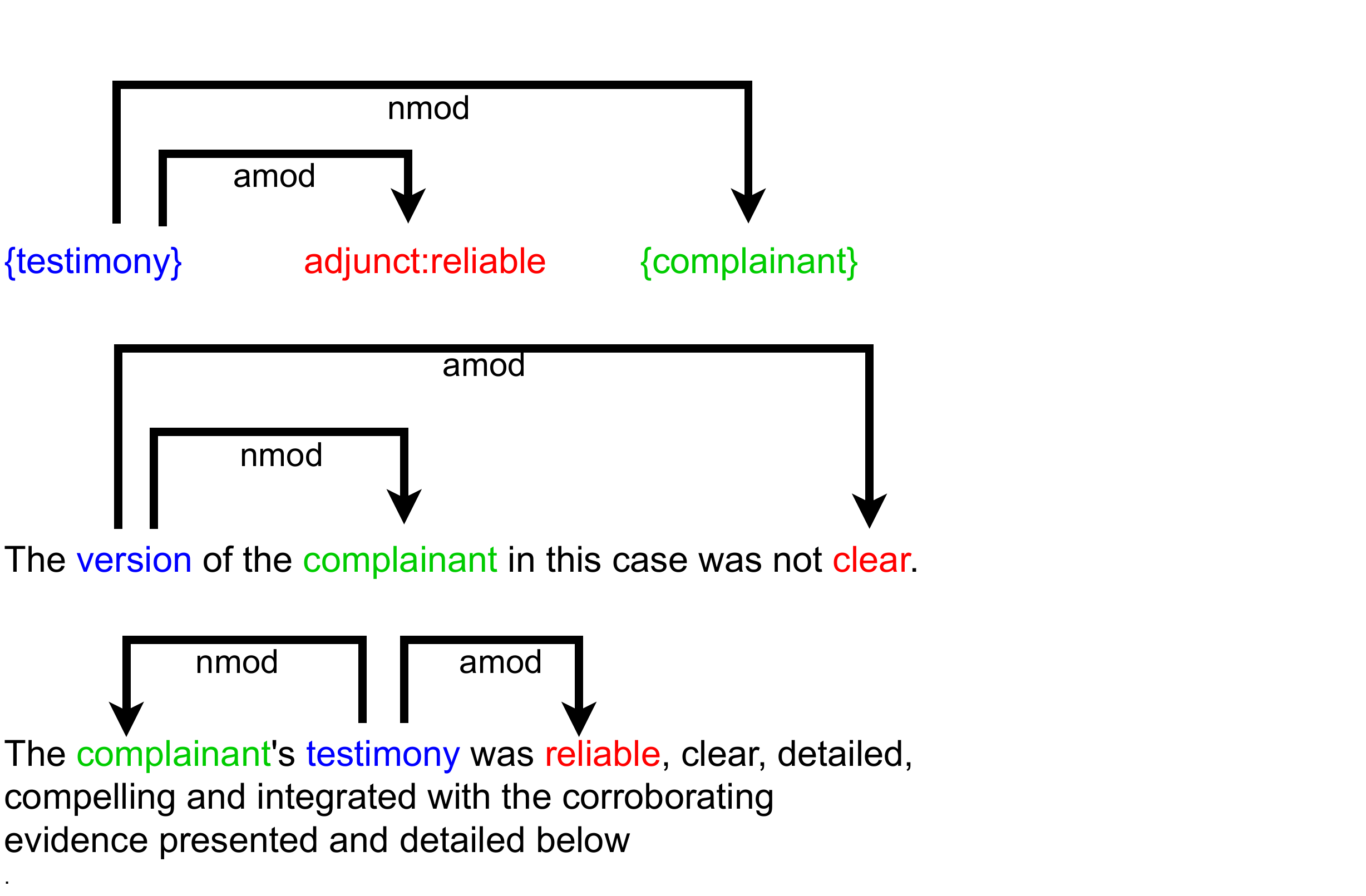}    
    \caption{
    An example of a syntactic pattern for query number 1 and sentences that align with this pattern. Words enclosed in brackets are predetermined lists of words and at least one of them must be included in the  sentence.}
    \label{figure:dt}
\end{figure}

In the first task, we take court documents and  aim to extract sentences that convey a judgment of the victim's credibility. To achieve this, we use  SPIKE to build syntactic trees of the sentences using queries related to the victim's testimony, as shown in Figure~\ref{figure:dt}. The figure illustrates the process of how SPIKE extracts and matches sentences with an example of a syntactic pattern from a query and the sentences that align with this pattern.
To formulate these queries, we started by manually identifying several seed sentences  that convey a judgment of the victim's credibility by the judge. Then, we extend and refine this seed by finding additional sentences and templates with the goal of reaching a coverage of at least one sentence per document. As a result, we constructed 19 queries, each producing a pattern of a sentence that deals with the credibility of the testimony.

Some of these queries differ  in the syntactic relations, while others result in  different syntactic trees. The syntactic trees generated by these queries will be provided in an Appendix upon publication.

\subsection{Modeling Credibility Classification}
\label{sec:classification}
In the second task, we classify the extracted sentences into predefined legal categories as defined in Section~\ref{section_4:legal_categories}.
To achieve this, we fine-tune one of the state-of-the-art Hebrew language models, AlephBERT~\cite{seker2022alephbert} for the credibility classification task by adding a task-specific output layer on top of the pre-trained model.

The data for the credibility classification task consisted of 200 tagged sentences. To ensure the reliability of our test results, we included only those sentences that were tagged by all annotators in our test set and achieved a high level of inter-annotator agreement. Although this resulted in a relatively small number of examples, this is done to ensure the accuracy of our test. The remaining data was divided into training and validation sets, with a split of 70\% for training and 30\% for validation. We trained the model for 30 epochs and averaged the results across 3 random seeds.



\section{Evaluation}
We start by reporting the results of sentence extraction (Section~\ref{sec:eval:sent}), including the high rate of extracted sentences and their consistency in labeling. In Section~\ref{sec:eval:classification}, we evaluate the fine-tuned model on credibility classification and find that it performs on-par with human agreement on the task,  demonstrating the potential to classify parts of the judgment using a few human labels, making it a valuable automatic measure. 

\subsection{Evaluating Sentence Extraction}
\label{sec:eval:sent}
\begin{table}[tb!]
\begin{tabular}{ll}
\hline

\textbf{Metric}          & \textbf{Score}   \\ \hline

Total Extracted Sentences           & 1662       \\ 
Documents with Multiple Sentences  & 43\%       \\ 
Documents with At Least One Sentence & 67\%       \\ 
Manually Verified Sentences    & 73\% \\
Sentences Corrected with Annotation Interface & 44\% \\
\hline
\end{tabular}
\caption{Results of the Sentence Extraction Task - This table presents the outcome of the sentence extraction process. 67\% of the documents yielded at least one sentence, and according to a manual evaluation of a sample of these sentences, at least 70\% were deemed correct based on the original task definition.
Additionally, a significant portion of 44\% of the sentences deemed incorrect during manual verification were easily corrected using the Annotation Interface}
\label{table:sentence_extraction}
\end{table}

The results for the first task are presented in Table~\ref{table:sentence_extraction}. Based on these we can draw several observations.  

\noindent \paragraph{High rate of extracted sentences.} 67\% of the judgments included at least one sentence relating to the complainant's testimony, which meets our expectation for every judgment to include such a sentence. In addition, 43\% of the judgments contained more than one extracted sentence. 

\noindent \paragraph{High consistency in high-level labels} 
In the 40\% of the documents where there were two sentences extracted, both sentences were tagged with the same granular label. Moreover, 70\% of these pairs had the same high-level label. These results suggest that the classification task could be performed at the document level for high-level labeling.

\noindent \paragraph{Diverse reasons for the \textsf{``not relevant''} label.} 
30\% of sentences extracted in the task were marked as \textsf{``not relevant''} by human taggers.
Further examination showed that these sentences were identified as not relevant for several reasons: 30\% of the sentences were labeled as not relevant as they did not fit the task definition of describing the judge's impression of the complainant, but instead described the impressions of other parties such as the defense or the prosecution; 20\% were general theoretical discussions by the judge, not specific to the case at hand; and 20\% were found to be relevant after review, but were mislabeled by the human taggers.

\subsection{Evaluating Credibility Classification}
\label{sec:eval:classification}
\begin{table}[tb!]
\begin{tabular}{lcc}
\hline
\textbf{Method}          & \textbf{\makecell{High Level\\ Accuracy}} & \textbf{\makecell{Granular Level\\ Accuracy}} \\ \hline
Random          & 6.2                          & 3.1                              \\
Majority        & 56.0                         & 44.0                             \\
IAA             & 72.8$\pm$4                         & 54.4$\pm$4                             \\
Finetuned  & \textbf{76.8}$\pm$5                & \textbf{63.7}$\pm$5                    \\ \hline
\end{tabular}
\caption{Comparing the performance of various methods in classifying credibility: A comparison of high-level accuracy and granular-level accuracy. The Finetune model has the highest accuracy, on par with human labeling as demonstrated by the inter-annotator agreement.}
\label{table:credibility_Classification_results}
\end{table}

Next, we evaluated the performance of the credibility classification task, as shown in Table~\ref{table:credibility_Classification_results}. Based on these results, we can make the following observation.

\noindent \paragraph{Our fine-tuned model is on par with human agreement in credibility classification.}
The results show that the fine-tuned model method achieved the highest accuracy, with 77\% at the high level and 64\% at the granular level. These results are on par with human labeling as demonstrated by the Inter-Annotator Agreement (IAA) scores. This highlights the potential of the fine-tuned model as a reliable alternative for credibility classification tasks. However, it also shows that the annotation process is noisy and subjective, and that future work can focus on improving the IAA before improving the model.

\section{Analysis}
This section provides a detailed analysis of our models for sentence extraction and credibility classification.
 
\begin{figure}[tb!]
\centering
\input{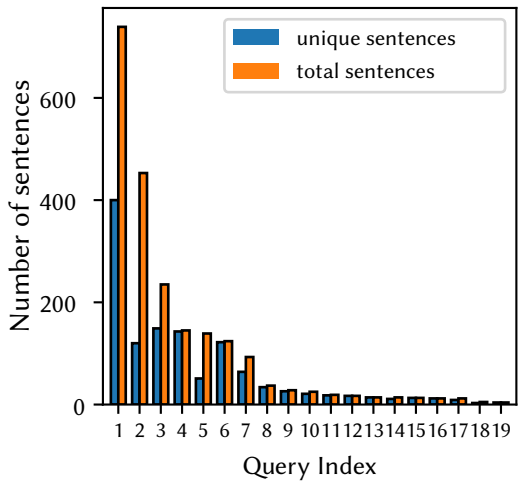}
\caption{Distribution of sentences matching each query, showing the number of unique sentences and total sentences matching each query.}
\label{figure:query_contribution}
\end{figure}

First, regarding the sentence extraction task,  Figure~\ref{figure:query_contribution} shows the distribution of the
sentences matching each query, as well as the number of unique
sentences, which are not captured by any other queries. 
As demonstrated in the Figure, the queries are quite diverse, each capturing different sentences.
 
Moving on to the credibility classification task, we draw additional conclusions from the model's results on this task and present several interesting phenomena that highlight the model's abilities and limitations in correctly recognizing the category. We provide examples to demonstrate each phenomenon clearly. To showcase the model's performance, we have only selected examples where the model is relatively confident in its answer.

\begin{table}[tb!]
\begin{tabular}{lcc}
\hline
\textbf{High Level}               & \textbf{Gold Data (\%)} & \textbf{Predicted Data (\%)} \\ \hline
Unequivocal Credibility & 49.7                   & 65.3                  \\Credible Because         & 23.2                  & 19.8                  \\Not Credible             & 18.6                  & 13.6                  \\Credible But             & 8.5                   & 1.3                   \\ \hline
\end{tabular}
\caption{Comparison of label distribution in training and test data with model's predictions. Results indicate similarity in overall distribution, with notable changes at the edges.}
\label{table:label_distribution}
\end{table}


\paragraph{Distribution similarity suggests potential for improvement in model performance.} In Table~\ref{table:label_distribution}, we compare the distributions of gold labels in the training data and predicted labels.
The KL divergence values between the gold and predicted labels are 0.23 for high-level categories and 0.36 for granular-level categories indicating a high degree of similarity. Nevertheless, the distributions do not match exactly, indicating potential limitations in the model. We suggest augmenting the training set with additional labeled data to increase the diversity of examples.

\paragraph{Our model struggles with double negation.} 
About $\sim$15\% of the model's errors occurred in instances of double negation~\cite{Wang2013EffectivenessEF}. As an example, the statement ``There is nothing in these claims that leads to the conclusion that the complainant's testimony is not credible'' was classified as \textsf{``Not credible''} by the model. However, the correct category is \textsf{``Unequivocal credibility''}. Our findings suggest that this error is likely caused by the presence of double negations in the sentence, which confuses the model. When the sentence is simplified and rephrased to contain only one negation or no negation at all, the model gives accurate results. For example, 
``These claims lead to the conclusion that the complainant's testimony is credible" was correctly classified  as \textsf{``Unequivocal credibility''}, and ``These claims lead to the conclusion that the complainant's testimony is not credible'' was correctly classified as \textsf{"Not credible''}.

\paragraph{Our model seems to be sensitive to paraphrasing.}
Our model has shown sensitivity to paraphrasing, with output that differs in both assigned label and level of confidence when input paraphrased statements into it. This challenge of achieving consistent predictions for paraphrased examples is a well-known issue in the NLP domain[16]. For example, the model classifies the statement ``The complainant's version was reliable, and one could get an impression of her innocence, despite her mature age'' as \textsf{``Credible because''}, 
but when the word \emph{``version''} is replaced with \emph{``testimony''}, the statement is classified as 
 \textsf{``Unequivocal credibility''}. This may indicate that the model is overfitting to the training data, which highlights the importance of selecting a diverse and representative training set.
 
\paragraph{Multi-word expressions seem to hinder the model's performance}
The model may sometimes err when it encounters multi-word expressions~\cite{sage:hal-03267497}~\cite{10.1007/3-540-45715-1_1}. For example, in the case of the testimony ``From the review of her testimony, the complainant turns out to be a witness for whom the truth is not a candle to her feet'' (a multi-word expression in Hebrew), the model classified it errornously as \textsf{``Unequivocal credibility''}.  However, when rephrased to ``From the review of her testimony, the complainant turns out to be a witness for whom the truth is not important'', the model correctly classified it as \textsf{``Not credible''}. 

\paragraph{Misclassification of abstract legal rules.} In some cases, the model misclassified a testimony as not credible when in fact the sentence included an abstract discussion of the legal rule rather than a concrete assertion regarding the credibility of the victim. For example, the model classified as \textsf{``Not credible''} the statements: ``No conviction is possible, even if the complainant's version is found to be true and has real weight'' and ``In a criminal trial it is not enough that the testimony of the complainant be more reasonable than the testimony of the accused, such a gap in evidence is sufficient for a decision in a civil trial'', even though these statements do not necessarily indicate a concrete decision in the credibility of the victim is the specific case at hand.


\section{Related Work}
Legal NLP has been getting increasing attention in recent years. Perhaps most relevant to our work is~\cite{wenger-etal-2021-automated} which focuses on the extraction of punishments in sentence decisions in cases of sexual assault in Israeli courts. It uses both rule-based and supervised models, demonstrating the effectiveness of both approaches. 
Also related, in their study on gender slant~\cite{ash2021gender} use NLP approaches, and specifically GloVe word embeddings, to identify gender attitudes in the language of published opinions in US circuit courts.

In a recent study on legal classification~\cite{chalkidis-etal-2019-large} explores multi-label text classification in the legal domain and releases 57K legislative documents. The results show that BIGRUs with label-wise attention and domain-specific word2vec and context-sensitive ELMo embeddings outperform other methods.
In their work,~\cite{Ashihara2020} propose the use of homophily networks to improve the topic modeling of legal case documents.~\cite{9207211} proposed NLP techniques for textual classification and achieved 90\% accuracy in categorizing 18 areas of law.
In~\cite{ Sexism_Judiciary}, 6.7 million case law documents are analyzed to determine gender bias in the judicial system, and alternative NLP approaches are proposed. 
While~\cite{inproceedings} compares the performance of deep learning with logistic regression and support vector machines in legal document review tasks, and shows that convolutional neural networks perform well with larger training datasets, BERT was compared to other machine learning text classification techniques in~\cite{GonzalezCarvajal2020ComparingBA} and was found to be superior. In~\cite{Bhattacharya2019IdentificationOR} the authors address the problem of automatically understanding rhetorical roles in Indian legal judgments. They use deep neural models and label sentences with human annotators, finding better performance compared to prior methods using handcrafted features. Using automated stance detection,~\cite{bergam-etal-2022-legal} analyzes US Supreme Court documents and proposes two ideology metrics. The study finds a correlation between the justices' language-based metrics and public opinion and presents the new legal stance detection task using the SC-stance dataset with competitive results using language adapters trained on legal documents.~\cite{inbook} discusses the application of multi-label classification algorithms to the EUR-Lex database of legal documents of the European Union, while~\cite{8511194} presents a comparative study of document classification using traditional machine learning and neural networks-based methods.


\section{Conclusion}
We formulated the first ontology for evaluating victim’s credibility focusing on sexual assaults cases in Hebrew in the Israeli justice system.
We created a dataset of annotated judicial statements labeling 
credibility in the Hebrew language, as well as a model that can extract credibility labels from court cases. The model we developed combines approaches of syntactic and latent structures to find sentences that convey the judge’s attitude towards the victim and classify them according to the credibility label set. No empirical legal research is without its limitations. We note that we can only assess judicial attitudes based on the text of the judgment, which may not fully reflect the judicial attitude. Ash et al. 

From a computational perspective, we find that our model performs on par with human agreement, suggesting that future work should focus on more consistent granular and high-level annotation before turning to additional model improvements. Furthermore, we thoroughly analyze the model's performance and identify several error modes which can be addressed in future work, including double negation which is a common cause for model confusion, and multi-word expressions.

From a legal perspective, we found that while most of the judicial statements found the victim's testimony credible, both positive and negative judicial attitudes echoed lingering rape myths. Existing legal scholarship on rape myths shows that one of the phenomenon's prevalent influences in assessing credibility of the survivor is closely related to the claim that the rape allegation are fake and motivated by a complainant's desire for vengeance. We show that even in the positive case, where the witness is found credible because she is not vengeful or where there is no fear of fake allegation (essentially two sides of the same coin), the presence of rape myth logic is evident.
Our findings also demonstrate the discrepancy between formal rules of evidence and the unique characteristics of 
sexual violence victimization.


\section{Acknowledgments}
This research was done in collaboration with Dr. Carmit Klar Chalamish, ARCCI and the Department of Criminology, Conflict Resolution, Management \& Negotiation Graduate Program, Bar-Ilan University, Israel. The research was supported by a grant from the Center for Interdisciplinary Data Science Research at the Hebrew University of Jerusalem (CIDR). We would like to thank Dana Epshtein, Itai Benshalom and Daniel Shwartstein for their valuable research assistance.

\bibliographystyle{ACM-Reference-Format}
\bibliography{sections/bib}

\end{document}